\documentclass[review]{elsarticle}

\usepackage{lineno,hyperref}
\usepackage{longtable,array}
\usepackage{booktabs,tabularx,makecell}
\usepackage{amsthm,amsmath,amssymb}
\usepackage{mathrsfs}
\usepackage[ruled]{algorithm2e}
\usepackage[section]{placeins}
\modulolinenumbers[5]
\journal{Knowledge-Based Systems}

\begin{document}

\begin{frontmatter}

\title{AutoDESS: AutoML Pipeline Generation of Classification with Dynamic Ensemble Strategy Selection}


\author[mymainaddress]{Yunpu Zhao}

\author[mysecondaryaddress]{Rui Zhang\corref{mycorrespondingauthor}}
\cortext[mycorrespondingauthor]{Corresponding author}
\ead{zhangrui@ict.ac.cn}

\author[mysecondaryaddress]{Xiaqing Li}

\address[mymainaddress]{University of Science and Techonology of China}
\address[mysecondaryaddress]{SKL of Computer Architecture, Institute of Computing Technology, Chinese Academy of Sciences}

\begin{abstract}
Automated machine learning has achieved remarkable technological developments in recent years, and building an automated machine learning pipeline is now an essential task. However, existing AutoML pipeline approaches adopt monotonous ensemble strategies across different machine learning classification tasks. They ignore the fact that no single ensemble strategy can perform best on all of the various classification tasks, and thus cannot meet the performance requirements in practice. To this end, we propose AutoDESS, an efficient AutoML approach to identifying the best ensemble strategy for machine learning classification tasks by combining different ensemble strategies' strengths. To our best knowledge, AutoDESS is the first trial in AutoML aiming at searching and optimizing ensemble strategies. In the comparison experiments, AutoDESS outperforms the state-of-the-art AutoML approaches while with the same CPU time in 42 classification datasets from the OpenML platform. In addition, ablation experiments also validate the effectiveness of AutoDESS.
\end{abstract}


\end{frontmatter}


\section{Introduction}

Machine Learning has achieved remarkable developments in a wide range of applications. Especially, ensemble learning plays a key role as the core approach for the model ensemble part of the AutoML framework. Two representative ensemble strategies are emerged in the last decades, including stacked generalization \cite{DBLP:journals/nn/Wolpert92} and static selection. Both of them typically adopt fixed ensemble strategies to exploit the high performance of the classification tasks. Concretely, auto-sklearn uses static selection, and TPOT \cite{DBLP:books/sp/19/OlsonM19}, AutoGluon \cite{DBLP:journals/corr/abs-2003-06505}, FLAML \cite{DBLP:conf/mlsys/0001WWZ21}, H2O-AutoML \cite{ledell2020h2o} are all based on stacked generalization, as shown in the upper part of Figure 1. While some approaches focus on the ensemble part of AutoML \cite{chen2018autostacker,wistuba2017automatic}, they are not the improvements from an ensemble strategy perspective but rather add complexity to an existing strategy (They build complex multi-layer stacked generalization automatically, but still stacked generalization). However, although delivering good performance, existing approaches just use one or two strategies in all of the different cases, which cannot consistently offer high performance in machine learning classification tasks.

The main reason is that each ensemble strategy has pros and cons which is known as No Free Lunch Theorem\cite{wolpert1997no}. For example, stacked generalization often delivers better performance than a single classifier. But due to the high complexity, the stacked generalization is easy to be overfitted, thus leading to great performance degradation, especially in the case of small datasets\cite{reid2009regularized}. Besides, it is quite hard for practitioners to choose the best for a given task in practice. The diverse research advance of ensemble learning brings difficulties for applying machine learning. Such diversity further increases the difficulty of the above choices. As such, selecting the optimal ensemble strategy is significant as well as challenging.

Therefore, to ease the choosing process, AutoML is proposed to identify the optimal ensemble strategy automatically. AutoML aims to reduce or even circumvent the involvement of human experts in the design process of machine learning\cite{he2021automl}\cite{zoller2021benchmark}. In recent years, researchers have attempted to automate the whole process to solve various machine learning tasks, which is called AutoML pipeline. AutoML pipeline is the same as the general machine learning, mainly divided into data preprocessing, feature engineering, and model selection. The choice of ensemble strategy should also be part of this. Unfortunately, existing AutoML pipeline generation approaches adopt monotonous ensemble strategies across different machine learning classification tasks. They ignore the fact that no single ensemble strategy can perform best on all of the various classification tasks, and thus cannot meet the performance requirements in practice.
\begin{figure}[htbp]
	\centering
	\includegraphics[height=6cm,width=12.5cm]{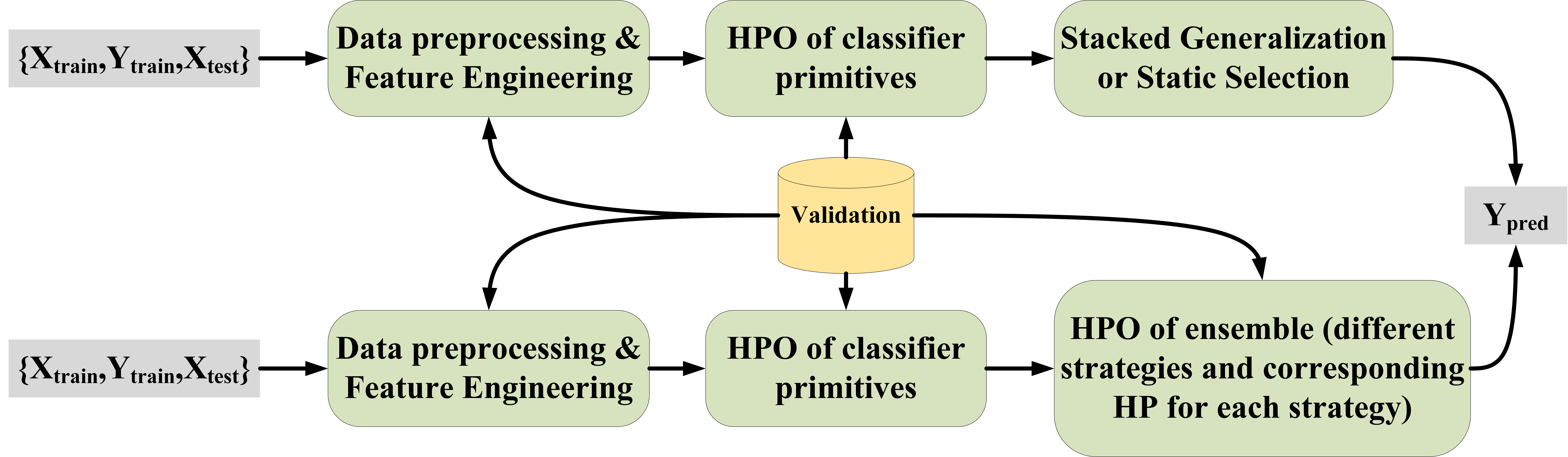}
	\caption{Differences between the current AutoML framework (upper) and our approach (below)}
\end{figure}

To this end, we aim to enrich and innovate the ensemble module of AutoML and thus propose AutoDESS, an efficient AutoML approach to identifying the best ensemble strategy for machine learning classification tasks by combing different ensemble strategies' strengths. As shown in the below part of Figure 1, AutoDESS can be divided into three modules. Firstly, it preprocessed the input data and feature engineering, where the algorithms used are selected (Data Prerprocesssing and Feature Engineering). Secondly, AutoDESS selects a subset of classifiers for the ensemble and chooses the appropriate strategy and hyperparameters (Hyperparameter Optimization for Classifiers and Ensemble Strategy). Specifically, AutoDESS deeply integrates Bayesian Optimization to conduct the optimization of the three modules. We compare AutoDESS on 42 public datasets of the OpenML platform with the current state-of-the-art approaches, including TPOT, Auto-sklearn, FLAML, and mljarsupervised\cite{DBLP:books/sp/19/OlsonM19}\cite{DBLP:conf/mlsys/0001WWZ21}\cite{feurer2015efficient}\cite{mljar}. The experimental result shows that AutoDESS outperforms the state-of-the-art AutoML approaches in most datasets while with the same CPU time (1800 seconds) in various metrics. Based on the average of the results, AutoDESS consistently outperforms existing approaches over most datasets, with f1 (accuracy) improvement of up to 3.4\% (0.6\%) against the best baseline. Also, our approach has the highest average ranking on all datasets.

To our best knowledge, AutoDESS is the first trial that searches and optimizes ensemble strategy in AutoML. The technical contribution of this paper is three-fold:
\begin{itemize}
\item This paper is the first work to introduce dynamic ensemble selection (DES)-related research advances into AutoML field.
\item We propose the AutoDESS approach. This method enables a significant improvement in model performance, especially for the imbalance problem, by constructing a rich pool of ensemble strategies and using Bayesian optimization technique.
\item We conduct comprehensive performance evaluation and analysis on different datasets, indicating that AutoDESS is able to identify the best or near-the-best strategy that delivers better performance while with the same CPU runtime, compared with existing approaches. We also did hypothesis testing and ablation experiments to demonstrate the effectiveness of the approach.
\end{itemize}

The rest of the paper is organized as follows. In Section 2, we briefly review the related work on the ensemble in AutoML pipeline, different ensemble strategy especially dynamic ensemble selection. Section 3 introduces some background concepts and motivation underlying our novel approach. Our proposed approach is described in detail in Section 4. Section 5 presents the experimental results including comparisons with some baselines, ablation study and some extra exploration of the results. Finally, Section 6 concludes the paper.

\section{Related Work}
In this section, we briefly review the Ensemble part in current AutoML Pipeline and the concept of dynamic ensemble selection (DES).
\subsection{Ensemble in AutoML Pipeline}
AutoML is a general term used to describe the process of automating the selection and optimization of ML algorithms and corresponding hyperparameters. As mentioned above, the AutoML pipeline is divided into data preprocessing, feature engineering, model selection, and model ensemble. AutoML Pipeline refers to automating the entire process of generating machine learning, rather than focusing on automating just one part of it. Existing approaches to the model ensemble are similar in their treatment, as shown in Table 1. It has to be admitted that stacked generalization is a powerful ensemble technique widely used in data science competitions. However, it also suffers from the tendency to overfit and the high complexity of training. Our approach offers not only a lightweight alternative to the stacked generalization but also more possibilities for the model ensemble.
\begin{table}[htbp]
	\centering
	\caption{Difference of current AutoML framework in ensemble part}
	\begin{tabular}{c c}
		\toprule
		AutoML Framework & Ensemble Method\\
		\midrule
		Auto-sklearn & Static Selection \\
		TPOT & Stacked Generalization \\
		H2O AutoML\cite{ledell2020h2o} & Stacked Generalization + Bagging \\
		AutoGluon & Stacked Generalization (Multi-Layer) \\
		FLAML & Stacked Generalization (Optional) \\
		mljarsupervised & Static Selection + Stacked Generalization\\
		\bottomrule
	\end{tabular}
\end{table}

\subsection{Ensemble Strategy}
In recent years, there has been much work focusing on advances in ensemble learning and multiple classifier systems\cite{britto2014dynamic}\cite{cruz2018dynamic}. In addition to the traditional stacked generalization, bagging and boosting methods, dynamic ensemble selection techniques have been developed. Dynamic ensemble selection (DES) is a subclass of model ensemble methods. DES techniques work by estimating the competence level of each classifier from a pool of classifiers. Only the most competent or an ensemble containing the most competent classifiers is selected to predict the label of a specific test sample. The technique's rationale is that no single classifier is an expert in classifying all samples, and it is natural to use different models to predict different kinds of samples.
\begin{figure}[htbp]
	\centering
	\includegraphics[height=6cm,width=12cm]{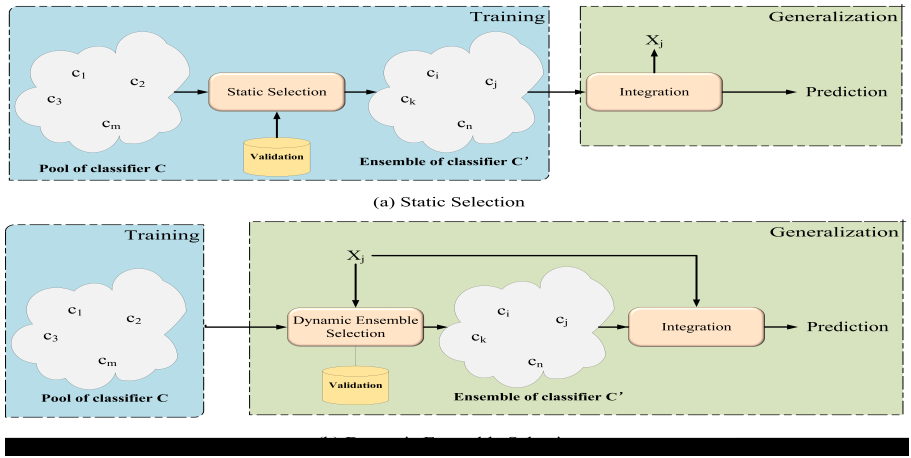}
	\caption{Differences between DES and static selection}
\end{figure}
In dynamic selections, the key is how to pick the most competitive classifier for any given sample. Usually, the capability of a classifier is estimated by the local region of the feature space of a given sample. This region can be defined by different methods, leading to various DES methods, such as applying the KNN technique to find the neighborhood of the query sample or using clustering approaches\cite{kuncheva2000clustering}\cite{soares2006using}. Then, the competence level of the base classifiers is estimated, considering only the samples belong to the same region as the query sample according to selection criteria such as accuracy, ranking, or probabilistic models\cite{woloszynski2012measure}.

However, the wide variety of DES methods poses great difficulties for machine learning researchers in selecting the appropriate DES strategy for a particular problem. In order to determine the best method, a lot of practice is generally needed to make a selection one by one, and the selection needs to be followed by tuning the hyperparameters. In our work, we consider DES with different methods as different ensemble strategies, and also consider traditional stacked generalization and static selection as independent ensemble strategies. Our work is essentially to select the appropriate strategies from a rich pool of ensemble strategies and find the corresponding hyperparameters and base classifier pool for ensembling. We should also mention here that \cite{cruz2020deslib} provides a great tool for using DES.
\section{Preliminaries}
This section will formalize the problem and describe the basics of AutoML pipeline generation and Bayesian Optimization. 

\subsection{Pipeline Creation Problem}
Let a triplet $ (g,\vec{A},\vec{\lambda}) $ define an ML pipeline with $ g $ a directed acyclic graph, $ \vec{A} $ a vector consisting of the selected algorithm for each node and $ \vec{\lambda} $ a vector comprising the hyperparameters of all selected algorithms. The pipeline is denoted as $ P_{g,\vec{A},\vec{\lambda}} $.

Let a trained pipeline $ P $ be given. Given a dataset $ D $ of size $ m $ and a loss metric $ L $, the performance $ \pi $ of $ P $ is calculated as:
\begin{equation}\label{eq1} \pi(P_{g,\vec{A},\vec{\lambda}},D)=\frac{1}{m}\sum_{i=1}^{m}L(\hat{y_i},y_i) 
\end{equation} with $ \hat{y_i} $ being the predicted output of $ P $.

Let a set of algorithm $ \mathcal{A} $ with an according domain of hyperparameters $ \Lambda $ and a set of valid pipeline structure $ G $ be given. Furthermore, let a dataset $ D $ be given. Then, the pipeline creation problem consists of finding pipeline structure together with a joint algorithm and hyperparameter selection that minimizes the loss:
\begin{equation}\label{eq2}
g^*,\vec{A^*},\vec{\lambda^*} \in {\underset{g \in G,\vec{A} \in \mathcal{A}^{|g|},\vec{\lambda} \in \Lambda}{{\arg\min}\, \pi({P_{g,\vec{A},\vec{\lambda}},D})}}
\end{equation}

As equation \ref{eq2} formulated, the pipeline creation problem is formulated as a black-box optimization problem. We can consider various optimization methods to solve such a problem. In this paper we use the classical Bayesian optimization using Gaussian processes.
\subsection{Bayesian Optimization}
Bayesian Optimization (BO) is an iterative algorithm that is popularly used for HPO problems. It determines the future evaluation point based on the previously-obtained results\cite{yang2020hyperparameter}. BO uses two components: the surrogate model and the acquisition function. The model is a regression model that fits all the currently-observed points into the objective function. After obtaining the predictive distribution of the probabilistic surrogate model, the acquisition function determines the usage of different points by balancing the trade-off between exploration and exploitation.

The reason for using BO as the optimization technique for our approach is that, on the one hand, compared to grid search and random search, it is much more scalable for higher dimension and is unlikely to end with local optima rather than global optima. On the other hand, compared to evolutionary optimization, BO does not require a great number of training cycles and are not noisy as well. Its scales well with utmost resource utilization, handling noisy data well exploiting non continuous spaces to attain global minima.

Then specifically for the proposed approach, we use the Gaussian process (GP) as the surrogate model in BO. Assuming that the function $ f $ with a mean $ \mu $ and a covariance $\sigma^2$ is a realization of a GP, the prediction follows a normal distribution:
\begin{equation}\label{eq3}
	p(y|x,D)=N(y|\hat{\mu},\hat{\sigma^2})
\end{equation}where D is the configuration space of hyperparameters, and $y=f(x)$is the evaluation result of each hyperparameter value $ x $. After a set of predicted data is obtained, the next point to be evaluated is selected from the BO-GP model's confidence intervals. The data from each new test is added to the sample record, and the BO-GP model is the rebuilt with the new information. This process is repeated until the end.
The loop of BO can be stated as:
For $ t=1:T $:
\begin{enumerate}
	\item Given observations$ (x_i,y_i=f(x_i))$ for $i=1:t$, build a probabilistic model for the objective. Integrate out all possible true functions, using Gaussian process regression.
	\item Optimize a cheap acquisition function $ u $ based on the posterior distribution for sampling the next point. $ x_{t+1}=\arg\min u(x)$ Exploit uncertainty to balance exploration against exploitation.
	\item Sample the next observation $y_{t+1}$ at $x_{t+1}$
\end{enumerate}

There are also many options of acquisition function $ u(x) $, such as the most commonly used expected improvement ( $ -EI(x)=-\mathbb{E}[f(x)-f(x_t^+)]$), lower confidence bound ($LCB(x)=\mu_GP(x)+\kappa\sigma_GP(x)$) or probability of improvement ($ -PI(x)=-P(f(x)\geq f(x_t^+)+\kappa)$), $\kappa$ is hyperparameter for controlling the trade-off between exploration and exploitation. In our approach, we use $gp_hedge$, a acquisition function that probabilistically choose one of the above three acquisition functions at every iteration. First, The gains $ g_i $ are initialized to zero. Then, at every iteration,
\begin{itemize}
	\item Each acquisition function is optimized independently to propose an candidate point $x_i$.
	\item Out of all these candidates, the next point $x_best$ is selected by $softmax(\eta g_i)$.
	\item After fitting the surrogate model with $(x_test,y_test)$, the gains are updated such that $g_i=\mu(X_i)$
\end{itemize}

AutoDESS aims to optimize the model's performance on the validation dataset to find the best or near-the-best ensemble strategy that offers high performance for a given machine learning classification tasks. In our problem, we have little knowledge about the objective performance model. There are two reasons for that. First, the relationship between the ensemble part and model performance is uncertain. Meanwhile, the strategies themselves are correlated in an unknown way, which increases such uncertainty further. Second, a limited amount of information supports the performance model since evaluations of an ensemble learning model require an amount of runtime, and so we can only afford a few of them. Therefore, we cannot give an accurate performance model based on the unknown relationship and limited information. Still, the information is sufficient identify a desirable strategy within a few tuning steps.

Therefore, instead of proposing an accurate performance model, we need a model that can guide AutoDESS in identifying the best strategy with the highest accuracy from the rest. Bayesian Optimization is an efficient approach to solve such an optimization problem, where the performance of the validation dataset is a black-box function that can be assumed through a few evaluations. By applying BO to it, AutoDESS is able to calculate the confidence interval according to samples taken from the training dataset. As the number of samples increases, the confidence interval area decreases, and the estimation of the performance of the validation dataset improves. In addition, AutoDESS can smartly suggest which strategy with its corresponding hyperparameters should be sample next to minimize the uncertainty in current modeling and more closer to the best one by using the predefined acquisition function as mentioned above.

\section{Proposed Method}

\subsection{Overview}
\begin{figure}[htbp]
	\centering
	\includegraphics[height=4cm,width=12cm]{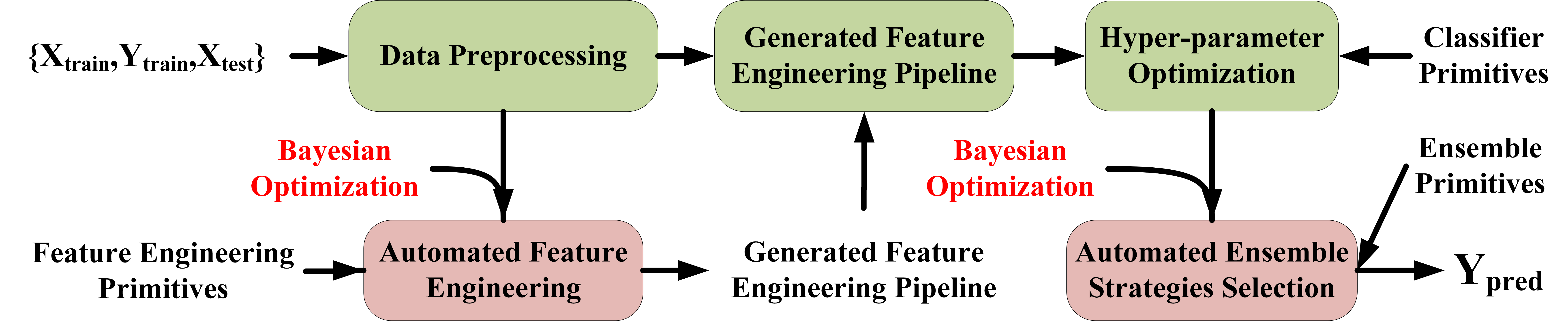}
	\caption{The overall framework of proposed AutoDESS}
\end{figure}
As shown in Figure 3, we have built an AutoML pipeline based on our approach. The input to the entire framework requires primitives sets on feature engineering, classifiers, and ensemble strategy, in addition to training sets and test data. The specific setting for primitives will be mentioned later.

There are three different optimization components in our framework. The first part is optimizing the feature engineering step, that is, finding the optimal combination of algorithms to perform the feature transformation, including matrix decomposition and dimensionality reduction. The second part is the optimization of the hyperparameters for every single model in the classifier primitives. After obtaining several tuned models, the third part is finding the optimal ensemble strategy and the corresponding hyperparameters, including which classifiers will be selected for ensemble learning. In the following, we will expand on the first and third parts, and the second part is not different from other AutoML approaches. We optimize feature engineering and ensemble separately because the dimensionality of the decision variables optimized by Bayesian optimization techniques should ideally not be too high; otherwise, it would reduce the efficiency of the search process.
\subsection{Automated Feature Engineering}
\begin{figure}[htbp]
	\centering
	\includegraphics[height=4cm,width=12cm]{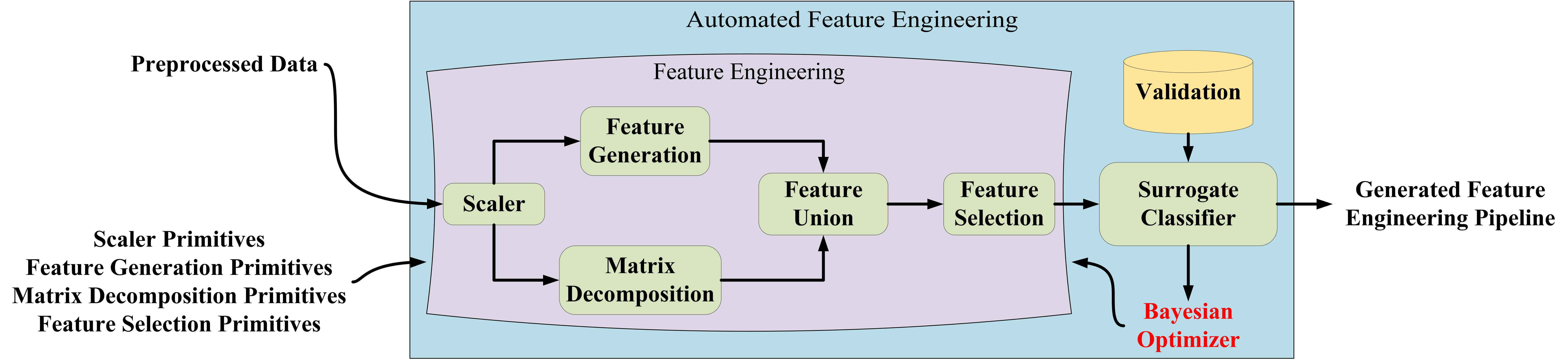}
	\caption{A simple design for automated feature engineering}
\end{figure}

The steps of automated feature engineering are shown in Figure 4. Before this, the data is preprocessed by the framework's preprocessing module, which focuses on missing value imputation and one-hot encoding of categorical features.

Feature Engineering may be the most important thing in a standard ML pipeline because it determines the upper bound of ML and algorithms can only approximate this limit. In our approach, we divide feature engineering into two main parts: feature transformation and feature selection\cite{motoda2002feature}\cite{dash1997feature}. The former is to transform the raw data into features that better represent the underlying problem of the prediction model, while the latter is to avoid the "curse of dimensionality" and reducing the computational complexity as well.

For the feature transformation, we further divide it into scaler, feature generation, matrix decomposition and feature union. The data is duplicated into two copies after the scaler. One copy of the data goes through a module for feature generation and the other for matrix decomposition. The two data are united and passed through the feature selection module to obtain the final dataset for training.

The next question is determining which algorithm to use for each module. Each of these modules has a corresponding selection of each module as a decision variable. In addition, we use a simple classifier as a surrogate model. The score of the surrogate model after the cross-validation on the training set can be considered the objective function of the optimization problem. This approach assumes that a promising feature engineering pipeline will already allow simple models to achieve relatively good results. It is worth noting that the meaning of the surrogate function referred to here is not the surrogate function used in Bayesian optimization. The former is used as a tool for generating values for the objective function and can be any machine learning model, while the latter is a surrogate model to approximate the objective function during optimization, usually using a Gaussian process model with Matern kernel.

Next, we introduce each module in more detail.
\paragraph{Scaler}
Scaler in our approach refers to a collective term for methods such as standardization and normalization of data. The data needs to obey certain rule after the scaler. For example, StandardScaler aims to standardize features by removing the mean and scaling to unit variance and MaxAbsScaler aims to scale each feature by its maximum absolute value. Also, RobustScaler can be used with the data with a lot of outliers that other methods are likely to not work very well. These are common methods used in machine learning task that can benefit learning process sometimes.
\paragraph{Feature Generation}
Sometimes feature generation can be called feature construction as well. It is a process that constructs new features from the basic feature space or raw data to enhance the robustness and generalizability of the model. Essentially, this is done to increase the representative ability of the original features\cite{he2021automl}. For example, PolynomialFeatures generates a new feature matrix consisting of all polynomial combinations of the features with degrees less than or equal to the specific degree (often set to 2). KBinsDiscretizer can bin continuous data into intervals. SplineTransformer can generate univariate B-spline bases of features\cite{perperoglou2019review}.
\paragraph{Matrix Decomposition}
We can call it signal composition either. The goal is extraction and separation of signal components from composite signals. Signal decomposition methods are closely related to classification of underlying features, which characterize the component to be separated\cite{Ohm2004}. In AutoDESS we consider it as a means of feature transformation and combine it with the feature matrix after feature generation. Some classical methods are included in this category such as principle component analysis, factor analysis and truncated singular value decomposition.
\paragraph{Feature Selection}
Feature selection builds a feature subset based on the original feature set by reducing irrelevant or redundant features. Feature selection can simplify the model that avoid over-fitting and improve model performance. This technique is especially important when the dataset has a large dimensionality. In our approach, the appropriate one is automatically selected among multiple feature selection methods. For example, removing features with low variance according to a threshold, removing all but a user-specified highest scoring percentage of feature, removing features according to false positive rate, false discovery rate or family wise error. In addition, recursive feature elimination is a very strong approach for feature selection even though it is more time-consuming.

\subsection{Automated Ensemble Strategy Selection}
\begin{figure}[htbp]
	\centering
	\includegraphics[height=4cm,width=12cm]{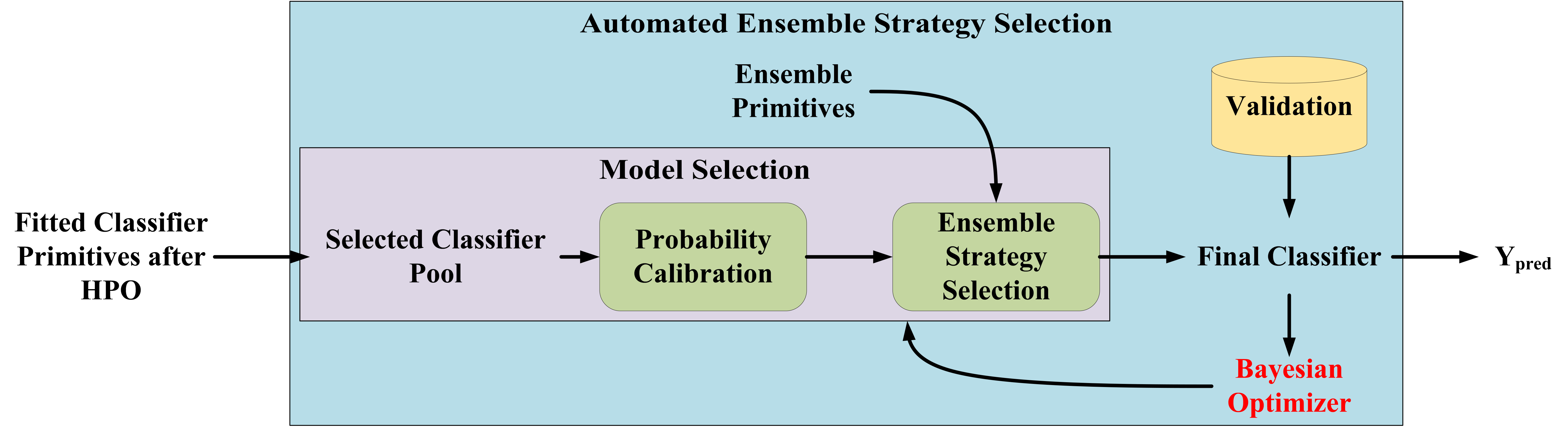}
	\caption{The design of Automated Ensemble Strategy Selection}
\end{figure}
The steps of automated ensemble strategy selection are shown in Figure 5. The input of this component is the fitted model primitives whose hyperparameters have been optimized. We then select a subset of classifier primitives based on the decision variables and perform a probability calibration for each classifier in the subset. The score of the ensemble model on the validation set is used as the objective function for the Bayesian optimization. In summary, there are three things to be optimized: what subset of classifier primitives we should select (selection of classifier pool), what strategy we should use to ensemble the model for this subset (Ensemble Strategy Selection), and how the hyperparameters of this strategy should be tuned (The Setting of Decision Variables and Probability Calibration). We will introduce each of these steps in detail.
\subsubsection{Decision Variable in Model Selection}
The decision variables indicate the optimization process of the component. Each classifier primitive corresponds to a Boolean variable in the subset selection: True corresponds to joining the subset and vice versa. The ensemble also has a variable whose value represents which strategy in the ensemble primitives is used. Finally, we set up decision variables regarding the internal hyperparameters of the ensemble strategy. Since most of our ensemble strategies belong to the DES technique, and most of the DES techniques use the KNN model to calculate the region of competence, $ k $ is a critical decision variable that we optimize. In addition, we set an extra Boolean variable which, when true, we use the dynamic frienemy pruning technique (DFP)\cite{oliveira2017online}, which will be mentioned later.
\subsubsection{Probability Calibration}
Classification models can be divided into probabilistic and non-probabilistic models: the former outputs the probability that the sample belongs to different classes, and the later gives a definite result through decision function. To use both non-probabilistic models (support vector machine, ridge classifier, etc.) and probability-based ensemble methods\cite{woloszynski2012measure}\cite{didaci2005study}\cite{woloszynski2011probabilistic}, we use the probability calibration technique \cite{niculescu2005predicting} to support probability prediction with non-probabilistic models.

In our approach, the probability calibration uses Platt's logistic model. Take a particular model on binary classification as an example. Let the output of a learning method be $ f(x) $. To get calibrated probabilities, pass the output through a sigmoid:
\begin{equation}\label{eq4}
	P(y=1|f)=\frac{1}{1+\exp(Af+B)}
\end{equation} where the parameters $ A $ and $ B $ are fitted using maximum likelihood estimation from a fitting training set $(f_i,y_i)$. Gradient descent is used to find $ A $ and $ B $ such that they are the solution to:
\begin{equation}\label{eq5}
	\arg\min_{A,B}{-\sum_iy_i\log(p_i)+(1-y_i)\log(1-p_i)}
\end{equation} where
\begin{equation}\label{eq6}
	p_i=\frac{1}{1+\exp(Af_i+B)}
\end{equation}
For multiclass predictions, we calibrate each class separately in a one-vs-rest fashion. When predicting probabilities, the calibrated probabilities for each class are predicted independently. As those probabilities do not necessarily sum to one, postprocessing is performed to normalize them.
\subsubsection{Ensemble Strategy}
Our ensemble strategy primitives fall into three main categories: static selection, dynamic classifier selection, and DES. The latter two can be collectively referred to as DES. 

For the static ensemble methods such as static selection and stacked generalization, they take the parameters of the classifier pool and are used to build the ensemble model. The static selection output the results of the ensemble with the majority voting rule. Stacked generalization is a method for combing estimators to reduce their biases. Its scheme uses a number of diverse models, each of which is trained on independent cross-validation examples of the original dataset. The outputs of these models, along with the original input data (optional), are then used as inputs to other generalizers, at a higher level in the stacking structure.

For the DES, in addition to the classifier pool, there is hyperparameter $k$ for DES methods because KNN is the key to determining the region of competence in the DES technique. The Boolean variable DFP is used to decide whether to use dynamic frienemy pruning. The details and types of DES methods are too much to be expanded here in this paper, as they can be found in \cite{britto2014dynamic}\cite{cruz2018dynamic}.
\subsubsection{Dynamic Frienemy Pruning}
DFP is a technique introduced in \cite{oliveira2017online} that can be used as a pre-selector to keep only based classifiers with decision boundaries crossing the region of competence of a test sample, helping to define more precisely the concept of "local competence" evaluation of base classifiers of the classification of the test sample. 

This method pre-selects the base classifiers by defining the frienemy. For the classification of a test sample, two samples are frienemies if:
\begin{enumerate}[(1)]
	\item two samples are located in the region of competence of the test sample.
	\item two samples have different classes. 
\end{enumerate}	
The flow of the algorithm for the DFP method is shown below, which is referenced from the original literature. In our approach, the DFP technique is controlled through a decision variable optimized in the component.

\begin{algorithm}[H]
	\SetAlgoNoLine
	\caption{DFP Method}
	\KwIn{pool of classifiers $C$, region of competence of the test sample $\psi$}
	\KwOut{pool of classifier after pruning $C_{pruned}$}
	$C_{pruned}$ $\leftarrow$ empty ensemble of classifier\\
	$F$ $\leftarrow$ all pairs of frienemies in $\psi$\\
	\For {$c_i$ in $C$} 
	{
		$\phi$ $\leftarrow$ samples in $\psi$ correctly classified by $c_i$ \\
		$F_i$ $\leftarrow$ frienemies in $\psi$
		\If{$|F_i|\geq1$}
		{
			$C_{pruned}$ $\leftarrow$ $C_{pruned} \cup c_i$
		}
	}
	\If{$|C_{pruned}|=0$}{$C_{pruned} \leftarrow C$}
\end{algorithm}

\section{Experiments}

Our experimental section is divided into four parts. In the first part, we present the relevant settings for the experiment. In the second part, we introduced the method compared to our proposed approach. After that, we offer the results and analyze them, and at last, we do additional ablation studies to validate and analyze the proposed approach.
\subsection{Experimental Settings}
\subsubsection{Dataset}
 We select 42 classical classification datasets from the OpenML platform. These datasets are diverse, with instances from 500 to 5000 and feature numbers ranging from 5 to 50, with half being binary and the rest being multiclass. The datasets include numerical and categorical and may contain missing values and sample imbalance issues. Table 2 shows the information of the 42 datasets.
 Here we need to introduce the "imbalance ratio" index. This is a well know index for measuring class balance:
 \begin{equation}
 	IR=\frac{N_{maj}}{N_{min}}
 \end{equation} where $N_{maj}$ is the sample size of the majority class and $N_{min}$ is the sample size of the minority class. If the problem has a high imbalance ratio, it can be considered more complex than a problem for which the ratio is smaller. There are also some other variants, and the most classic definition is used here\cite{DBLP:journals/csur/LorenaGLSH19}.
 \begin{table}[htbp]
 	\centering
 	\caption{Datasets description. Column $N$ is the number of instances in the dataset, column $Feat$ is the number of features, column $k$ is the number of classes, column IR is the imbalance ratio of the dataset.}
 	\renewcommand{\arraystretch}{0.7}
 	\begin{tabular}{c c c c c} 
 		\toprule
 		Dataset & N & Feat & k & IR \\
 		\midrule
 		analcatdata-authorship & 841 & 71 & 4 & 5.76 \\
 		analcatdata-dmft & 797 & 5 & 6 & 1.26 \\
 		autoUniv-au6-750 & 750 & 41 & 8 & 2.89 \\
 		autoUniv-au7-1100 & 1100 & 13 & 5 & 1.99 \\
 		balance-scale & 625 & 5 & 3 & 5.88 \\
 		banknote-authentification & 1372 & 5 & 2 & 1.25 \\
 		blood-transfusion-service-center & 748 & 5 & 2 & 3.20 \\
 		breast-w & 699 & 10 & 2 & 1.90 \\
 		cardiotocography & 2126 & 36 & 10 & 10.92 \\
 		climate-model-simulation-crashes & 540 & 21 & 2 & 10.74 \\
 		cmc & 1473 & 10 & 3 & 1.89 \\
 		credit-g & 1000 & 21 & 2 & 2.33 \\
 		diabetes & 768 & 9 & 2 & 1.87 \\
 		eucalyptus & 736 & 20 & 5 & 2.04 \\
 		fri-c1-1000-10 & 1000 & 11 & 2 & 1.29 \\
 		fri-c2-1000-10 & 1000 & 11 & 2 & 1.38 \\
 		ilpd & 583 & 11 & 2 & 2.49 \\
 		kc1 & 2109 & 22 & 2 & 5.47 \\
 		mfeat-fourier & 2000 & 77 & 10 & 1.00 \\
 		mfeat-karhunen & 2000 & 65 & 10 & 1.00 \\
 		mfeat-morphological & 2000 & 7 & 10 & 1.00 \\
 		mfeat-zernike & 2000 & 48 & 10 & 1.00 \\
 		monks-problem-2 & 601 & 7 & 2 & 1.92 \\
 		pbcseq & 1945 & 19 & 2 & 1.00 \\
 		pc1 & 1109 & 22 & 2 & 13.4 \\
 		pc3 & 1563 & 38 & 2 & 8.77 \\
 		pc4 & 1458 & 38 & 2 & 7.19 \\
 		phoneme & 5404 & 6 & 2 & 2.41 \\
 		qsar-biodeg & 1055 & 42 & 2 & 1.96 \\
 		quake & 2178 & 4 & 2 & 1.25 \\
 		segment & 2310 & 20 & 7 & 1 \\
 		steel-plates-fault & 1941 & 28 & 7 & 12.24 \\
 		stock & 950 & 10 & 2 & 1.06 \\
 		tokyo1 & 959 & 45 & 2 & 1.77 \\
 		vehicle & 846 & 19 & 4 & 1.10 \\
 		volcanoes-a4 & 1515 & 4 & 5 & 47.07 \\
 		vowel & 990 & 14 & 2 & 10 \\
 		wdbc & 569 & 31 & 2 & 1.68 \\
 		wilt & 4839 & 6 & 2 & 17.54 \\
 		yeast & 1484 & 9 & 10 & 92.6 \\
 		credit-approval & 690 & 16 & 2 & 1.25 \\
 		boston & 506 & 14 & 2 & 1.42 \\
 		\bottomrule
 	\end{tabular}
 \end{table}
\subsubsection{Evaluation Metrics}
In addition to the most commonly used accuracy score, we add additional comparison tests using the $F_1$ measure. $F_1$ is the harmonic average of recall and precision, which can consider the accuracy and recall rate of majority and minority classes simultaneously. Under the experimental conditions in this paper, $F_1$ is a more appropriate evaluation indicator.
\begin{equation}\label{eq7}
	Precision=\frac{TP}{TP+FP}   Recall=\frac{TP}{TP+FN}
\end{equation}
\begin{equation}\label{eq8}
	F_1=\frac{2\times Precision\times Recall}{Precision+Recall}
\end{equation}
\subsubsection{Primitives Configuration}
This section shows the specific setup of the proposed method in the experiment.

\paragraph{Data Preprocessing \& Feature Engineering }
Our data preprocessing module automatically one-hot encodes all categorical features and imputes the missing value using the average. The table below shows our primitives for each module in the automated feature engineering. The surrogate classifier used in automated feature engineering is an extra-tree classifier. The process is cross-validated on the training set with a 5-fold.
\begin{table}[htbp]
	\centering
	\caption{Primitives used in automated feature engineering}	
	\renewcommand\tabularxcolumn[1]{m{#1}}
	\newcolumntype{C}{>{\centering\arraybackslash}X}
	\begin{tabularx}{350pt}{|c|C|}
		\hline
		Module & Primitives \\
		\hline
		Scaler & StandardScaler MaxAbsScaler RobustScaler Normalizer \\
		\hline
		Feature Generation & PolynomialFeature KBinsDiscretizer SplineTransformer Nystroem RBFSampler \\
		\hline
		Matrix Decomposition & FastICA IncrementalICA PCA SparsePCA TruncatedSVD Factoranalysis \\
		\hline
		Feature Selection & SelectFwe SelectFdr SelectFpr SelectPercentile VarianceThreshold RFE \\
		\hline
	\end{tabularx}
\end{table}
\paragraph{Classifier HPO \& Automated Ensemble Strategy Selection}
We set 18 different classifiers and perform HPO before making ensemble selection. It is worth noting here that the HPO can be shorter in terms of search time because the performance of a single model is not critical in our approach. We have chosen 23 differnt ensemble strategies as primitives, as shown below.
\begin{table}[htbp]
	\centering
	\caption{Primitives used in automated ensemble strategy selection}	
	\renewcommand\tabularxcolumn[1]{m{#1}}
	\newcolumntype{C}{>{\centering\arraybackslash}X}
	\begin{tabularx}{350pt}{|c|C|}
		\hline
		Classifier Primitives & HistGradientBoostingClassifier RidgeClassifier GaussianNB BernoulliNB BaggingClassifier DecisionTreeClassifier ExtraTreesClassifier RandomForestClassifier GradientBoostingClassifier KNeighborsClassifier LinearSVC SGDClassifier LogisticRegression Perceptron MLPClassifier LGBMClassifier PassiveAggressiveClassifier RUSBoostClassifier\\
		\hline
		Ensemble Strategies Primitives & SingleBest StackedGeneralization StaticSelection LCA MCB OLA Rank DESClustering DESKNN DESMI KNORAE KNORAU KNOP METADES DESKL Exponential Logarithmic RRC MinimumDifference APriori APosteriori \\
		\hline
	\end{tabularx}
\end{table}
For more information about primitives, please refer to the relevant open-source library\footnote{https://scikit-learn.org/stable/}\footnote{https://deslib.readthedocs.io/en/latest/}\footnote{https://imbalanced-ensemble.readthedocs.io/en/latest/index.html}.
\subsection{Compared Methods}
It is important to stress that we run all baselines for 1800 seconds on the same machine to ensure fairness. However, for mljarsupervised, it terminates early. Our search space was far smaller compared to the baselines. Under such circumstances, the experiments is enough to demonstrate the effectiveness of the proposed approach. All baselines are run using the AutoML-benchmark designed by OpenML\footnote{https://openml.github.io/automlbenchmark/}.
\paragraph{Auto-sklearn}
Auto-sklearn\cite{feurer2015efficient} is a tool for building machine learning pipelines. The pipeline all have a fixed structure: a fixed set of data cleaning steps including optional categorical encoding, imputation, removing variables with low variance, and optional scaling is executed. Then, an optional preprocessing and mandatory model algorithm are selected and tuned via the SMAC optimization method\cite{JMLR:v23:21-0888}.
\paragraph{TPOT}
TPOT\cite{DBLP:books/sp/19/OlsonM19} is a framework for building and tuning arbitrary machine learning pipelines. It uses genetic programming to construct flexible pipelines and selects an algorithm in each pipeline stage. Regarding HPO, TPOT can only handle categorical parameters, so all continuous hyperparameters have to be discretized. TPOT's ability to create arbitrary complex pipelines makes it prone to overfitting. TPOT optimizes a combination of high performance and low pipeline complexity. Therefore, pipelines are selected from the Pareto front using a multi-objective selection strategy. But, this approach is quite time-consuming.
\paragraph{FLAML}
FLAML\cite{DBLP:conf/mlsys/0001WWZ21} leverages the search space structure to choose a search order optimized for both cost and error. It iteratively decides the learner, hyperparameters, sample size, and resampling strategy while leveraging their pound impact on both cost and error as the search proceeds. The search gradually moves from cheap trials and accurate models to expensive trials and accurate  models. It is designed for robustly adapting to an ad-hoc dataset out of the box without relying on expensive preparation such as meta-learning.
\paragraph{MLjar-supervised}
mljar-supervised\cite{mljar} is an AutoML python package that works with tabular data. It uses many algorithms and can compute ensembles based on the greedy algorithm. It is a very lightweight AutoML framework, and we select it as a baseline.

\subsection{Results}
For each dataset, we use ten different random seeds for the training-test split and ten different random seeds for each split for testing, which circumvents both split and test randomness. Meanwhile, we performed hypothesis testing on the differences of the five algorithms.
\subsubsection{AutoDES Performance Analysis}
The experimental results for the average accuracy score and $F_1$ score on all datasets are shown in Figure 6. Figures 7 and 8 show the results of each method of testing on each of our datasets. We use the box-plot to visualize the overall results, and the box extends from the first quartile to the third quartile of the data, with a line at the median. In the scatterplot, the scatter of our approach is adjusted to be the largest to highlight our ranking on each dataset. Table 5 shows the specific values of the results while we report the running time as well. In the table, we bolded the best result among all baselines, and underlined the second best. We also show the raw data for the experiment in Table 6.
When using equal time budgets, AutoDESS clearly outperforms every competitor on the majority of the datasets. It is worth noting that the TPOT framework made errors when processing the four datasets and mljar-supervised made on one dataset, so the scores in the scatterplot are zero. We did not consider these failed datasets when calculating the average score. The results also show that our accuracy is slightly better than baselines. However, the $F_1$ scores are significantly higher than baselines, indicating the relative advantage of our method when dealing with imbalanced datasets.

To verify the correctness of the conclusions, we performed hypothesis testing of the results, as shown in Table \ref{H1} and Table \ref{H2}. We used the Wilcoxon signed-rank test\cite{woolson2007wilcoxon} as the hypothesis test for this paper because the data obtained from the 42 datasets did not satisfy a normal distribution and each dataset existed in a paired form among the five algorithms. The results show that our method is significantly better than baselines, except that there is no significant difference between Auto-sklearn and our method in terms of accuracy, and the remaining hypothesis tests reject the original hypothesis.
\begin{table}[htbp]
	\centering
	\caption{Quantification of results}
	\renewcommand\tabularxcolumn[1]{m{#1}}
	\newcolumntype{C}{>{\centering\arraybackslash}X}
	\begin{tabularx}{350pt}{cCCCCC}
		\toprule
		&TPOT&Auto-sklearn&FLAML&mljar&AutoDESS\\
		\midrule
		Average ACC&0.82433&\underline{0.83260}&0.83053&0.81803&\textbf{0.83804}\\
		Average $F_1$ &0.71351&0.73663&\underline{0.75107}&0.74244&\textbf{0.77722}\\
		Average ACC Rank&2.47368&\underline{2.38095}&2.78571&3.21951&\textbf{2.16667}\\
		Average $F_1$ Rank&2.88095&\underline{2.71429}&2.85714&3.19048&\textbf{2.04762}\\
		Average Time (s)&1793&1830&1802&143&998\\
		\bottomrule
	\end{tabularx}
\end{table}
\begin{figure}[htbp]
	\centering
	\includegraphics[height=7cm,width=12cm]{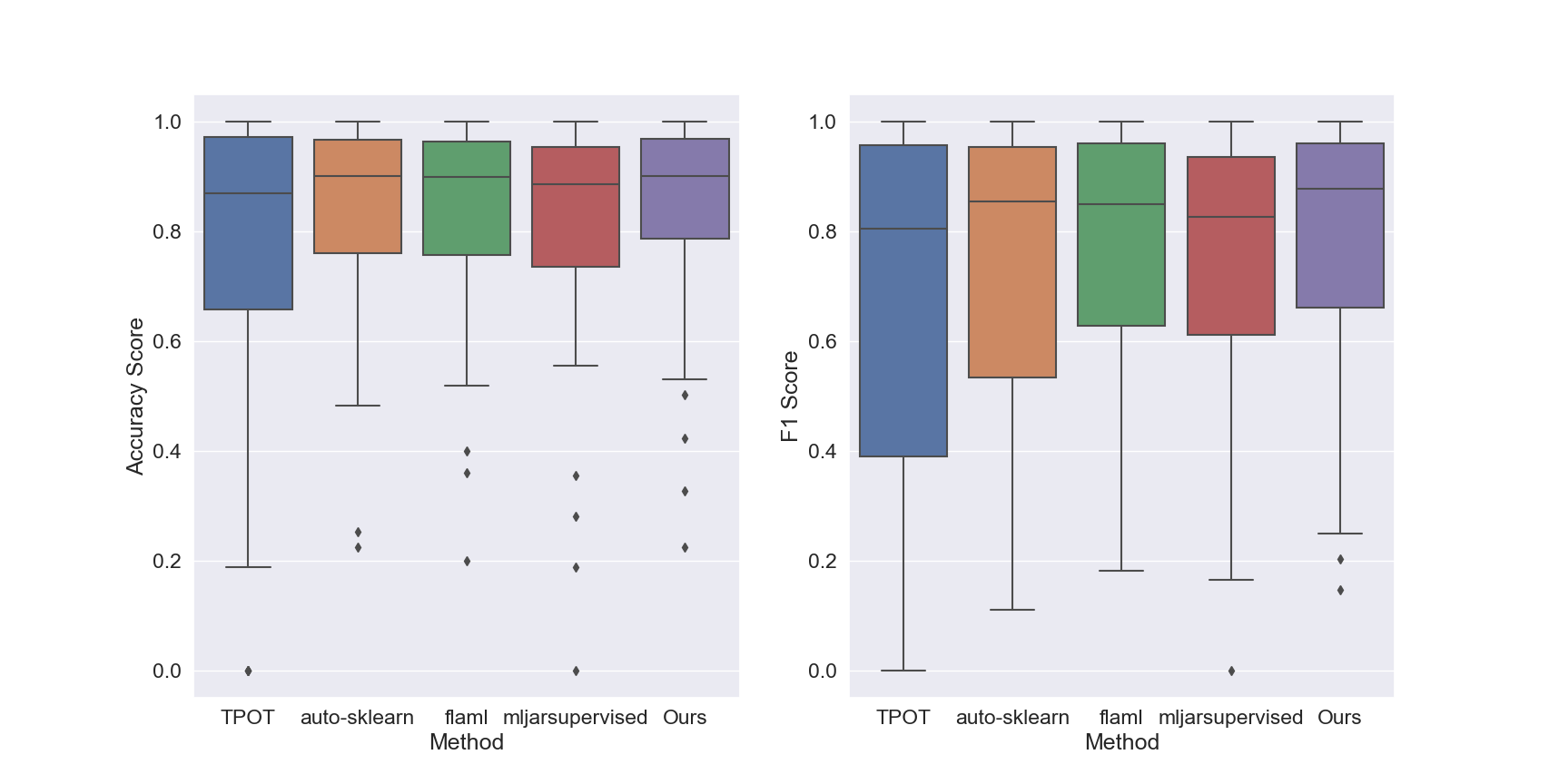}
	\caption{The boxplot of the comparison result}
\end{figure}
\begin{figure}[htbp]
	\centering
	\includegraphics[height=9cm,width=14cm]{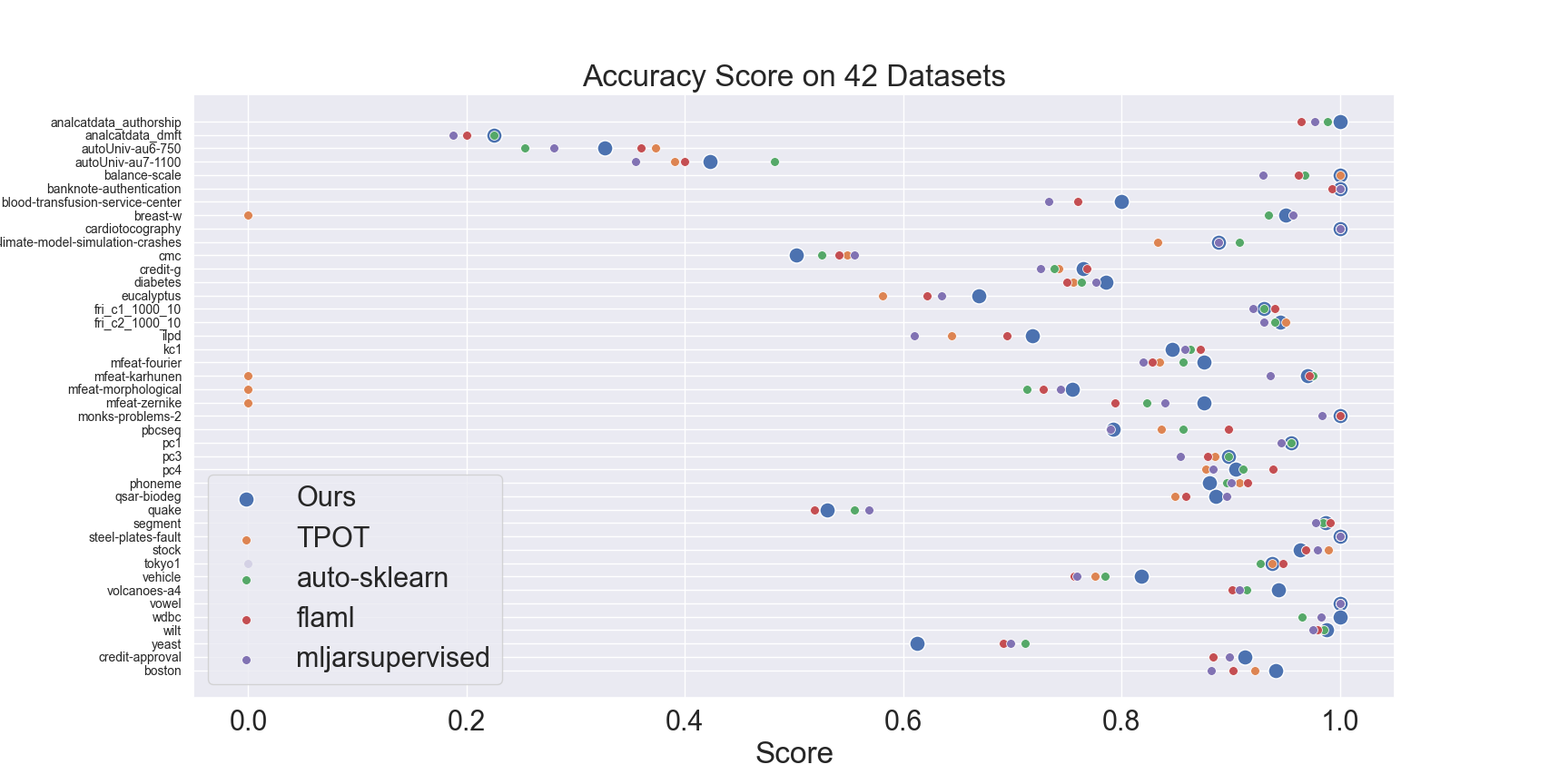}
	\caption{Accuracy score on 42 datasets}
\end{figure}
\begin{figure}[htbp]
	\centering
	\includegraphics[height=9cm,width=14cm]{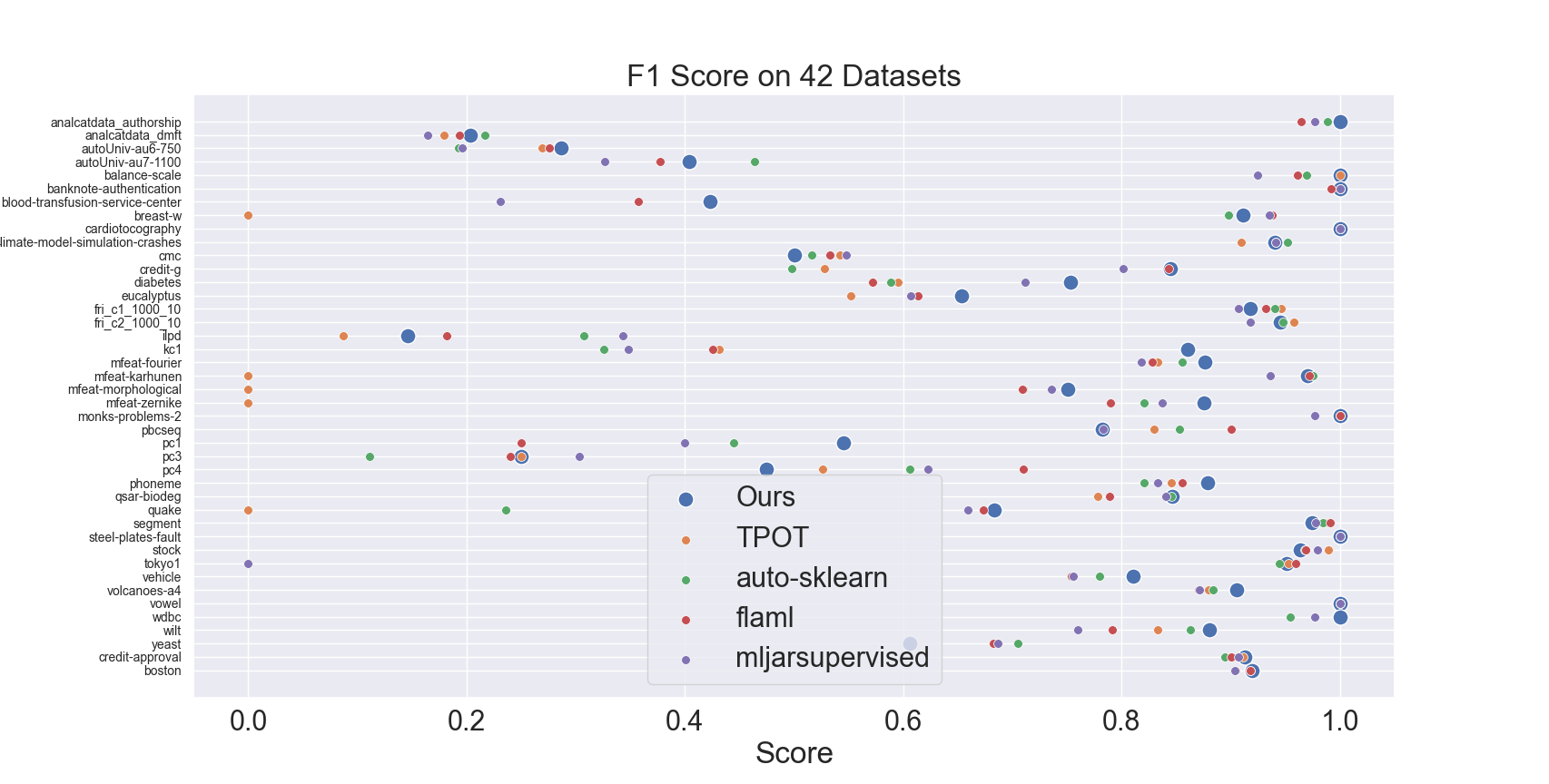}
	\caption{$F_1$ score on 42 datasets}
\end{figure}

\begin{table}[htbp]
	\centering
	\caption{The raw data of the comparison experiment}
	\resizebox{\textwidth}{80mm}{
	\begin{tabular}{ccccccccccc} 
		\toprule
		dataset\_name                    & mljarsupervised\_acc & mljarsupervised\_f1 & TPOT\_acc & TPOT\_f1 & auto-sklearn\_acc & auto-sklearn\_f1 & flaml\_acc & flaml\_f1 & Ours\_acc & Ours\_f1  \\ 
		\hline
		analcatdata\_authorship          & 0.976471             & 0.976421            & 0.976471  & 0.976421 & 0.988235          & 0.988226         & 0.964706   & 0.96429   & 1         & 1         \\
		analcatdata\_dmft                & 0.1875               & 0.164157            & 0.1875    & 0.179621 & 0.225             & 0.216597         & 0.2        & 0.19376   & 0.225     & 0.203125  \\
		autoUniv-au6-750                 & 0.28                 & 0.196211            & 0.373333  & 0.268848 & 0.253333          & 0.192689         & 0.36       & 0.275965  & 0.326667  & 0.286667  \\
		autoUniv-au7-1100                & 0.354545             & 0.326684            & 0.390909  & 0.376153 & 0.481818          & 0.463337         & 0.4        & 0.376962  & 0.422727  & 0.403653  \\
		balance-scale                    & 0.929712             & 0.9241              & 1         & 1        & 0.968051          & 0.969123         & 0.961661   & 0.960817  & 1         & 1         \\
		banknote-authentication          & 1                    & 1                   & 1         & 1        & 1                 & 1                & 0.992754   & 0.99187   & 1         & 1         \\
		blood-transfusion-service-center & 0.733333             & 0.230769            & 0.76      & 0.357143 & 0.76              & 0.357143         & 0.76       & 0.357143  & 0.8       & 0.423077  \\
		breast-w                         & 0.957143             & 0.935622            & 0         & 0        & 0.934286          & 0.897778         & 0.957143   & 0.937759  & 0.95      & 0.911111  \\
		cardiotocography                 & 1                    & 1                   & 1         & 1        & 1                 & 1                & 1          & 1         & 1         & 1         \\
		climate-model-simulation-crashes & 0.888889             & 0.941176            & 0.833333  & 0.909091 & 0.907407          & 0.951456         & 0.888889   & 0.941176  & 0.888889  & 0.940594  \\
		cmc                              & 0.554953             & 0.547815            & 0.548168  & 0.542057 & 0.525102          & 0.516149         & 0.541384   & 0.532524  & 0.501695  & 0.499904  \\
		credit-g                         & 0.726                & 0.801161            & 0.742     & 0.527473 & 0.738             & 0.498084         & 0.768      & 0.843243  & 0.765     & 0.844884  \\
		diabetes                         & 0.776042             & 0.711409            & 0.755208  & 0.594828 & 0.763021          & 0.588235         & 0.75       & 0.571429  & 0.785714  & 0.753247  \\
		eucalyptus                       & 0.635135             & 0.607096            & 0.581081  & 0.551666 & 0.635135          & 0.607241         & 0.621622   & 0.613512  & 0.668919  & 0.653288  \\
		fri\_c1\_1000\_10                & 0.92                 & 0.906977            & 0.94      & 0.946429 & 0.93              & 0.940171         & 0.94       & 0.931818  & 0.93      & 0.917647  \\
		fri\_c2\_1000\_10                & 0.93                 & 0.917647            & 0.95      & 0.957983 & 0.94              & 0.947368         & 0.93       & 0.917647  & 0.945     & 0.945     \\
		ilpd                             & 0.610169             & 0.342857            & 0.644068  & 0.086957 & 0.694915          & 0.307692         & 0.694915   & 0.181818  & 0.717949  & 0.146341  \\
		kc1                              & 0.85782              & 0.347826            & 0.862559  & 0.431373 & 0.862559          & 0.325581         & 0.872038   & 0.425532  & 0.845972  & 0.86019   \\
		mfeat-fourier                    & 0.82                 & 0.817768            & 0.835     & 0.833316 & 0.856             & 0.855636         & 0.828      & 0.827709  & 0.875     & 0.875862  \\
		mfeat-karhunen                   & 0.936                & 0.93613             & 0         & 0        & 0.975             & 0.975077         & 0.972      & 0.971972  & 0.97      & 0.970088  \\
		mfeat-morphological              & 0.744                & 0.735372            & 0         & 0        & 0.713             & 0.709908         & 0.728      & 0.709041  & 0.755     & 0.750774  \\
		mfeat-zernike                    & 0.84                 & 0.837109            & 0         & 0        & 0.823             & 0.820441         & 0.794      & 0.789394  & 0.875     & 0.875068  \\
		monks-problems-2                 & 0.983607             & 0.976744            & 1         & 1        & 1                 & 1                & 1          & 1         & 1         & 1         \\
		pbcseq                           & 0.789744             & 0.783069            & 0.835897  & 0.829787 & 0.85641           & 0.852632         & 0.897436   & 0.9       & 0.791774  & 0.782609  \\
		pc1                              & 0.945946             & 0.4                 & 0.954955  & 0.444444 & 0.954955          & 0.444444         & 0.945946   & 0.25      & 0.954955  & 0.545455  \\
		pc3                              & 0.853503             & 0.30303             & 0.88535   & 0.25     & 0.898089          & 0.111111         & 0.878981   & 0.24      & 0.897764  & 0.25      \\
		pc4                              & 0.883562             & 0.622222            & 0.876712  & 0.526316 & 0.910959          & 0.606061         & 0.938356   & 0.709677  & 0.90411   & 0.474576  \\
		phoneme                          & 0.900185             & 0.833333            & 0.907579  & 0.845679 & 0.896488          & 0.820513         & 0.914972   & 0.855346  & 0.880666  & 0.878816  \\
		qsar-biodeg                      & 0.896226             & 0.84058             & 0.849057  & 0.777778 & 0.896226          & 0.84507          & 0.858491   & 0.788732  & 0.886256  & 0.846343  \\
		quake                            & 0.568807             & 0.65942             & 0.555046  & 0        & 0.555046          & 0.23622          & 0.518349   & 0.672897  & 0.529817  & 0.682853  \\
		segment                          & 0.977489             & 0.977536            & 0.986147  & 0.986145 & 0.984416          & 0.984417         & 0.991342   & 0.991336  & 0.987013  & 0.974124  \\
		steel-plates-fault               & 1                    & 1                   & 1         & 1        & 1                 & 1                & 1          & 1         & 1         & 1         \\
		stock                            & 0.978947             & 0.979167            & 0.989474  & 0.989247 & 0.968421          & 0.967742         & 0.968421   & 0.968421  & 0.963158  & 0.963158  \\
		tokyo1                           & 0                    & 0                   & 0.9375    & 0.952381 & 0.927083          & 0.944            & 0.947917   & 0.95935   & 0.9375    & 0.95082   \\
		vehicle                          & 0.758865             & 0.755618            & 0.775414  & 0.77971  & 0.78487           & 0.779419         & 0.756501   & 0.753764  & 0.817647  & 0.810864  \\
		volcanoes-a4                     & 0.907895             & 0.871393            & 0.907895  & 0.87944  & 0.914474          & 0.883915         & 0.901316   & 0.870751  & 0.943894  & 0.905206  \\
		vowel                            & 1                    & 1                   & 1         & 1        & 1                 & 1                & 1          & 1         & 1         & 1         \\
		wdbc                             & 0.982456             & 0.976744            & 0.982456  & 0.976744 & 0.964912          & 0.954545         & 0.982456   & 0.976744  & 1         & 1         \\
		wilt                             & 0.975207             & 0.76                & 0.983471  & 0.833333 & 0.985537          & 0.862745         & 0.979339   & 0.791667  & 0.987603  & 0.88      \\
		yeast                            & 0.697987             & 0.686872            & 0.697987  & 0.686777 & 0.711409          & 0.704862         & 0.691275   & 0.68277   & 0.612795  & 0.606061  \\
		credit-approval                  & 0.898551             & 0.906667            & 0.898551  & 0.911392 & 0.884058          & 0.894737         & 0.884058   & 0.9       & 0.913043  & 0.912674  \\
		boston                           & 0.882353             & 0.903226            & 0.921569  & 0.904762 & 0.901961          & 0.918033         & 0.901961   & 0.918033  & 0.941176  & 0.919118  \\
		\bottomrule
	\end{tabular}}
\end{table}

\begin{table}[htbp]
	\centering
	\caption{Hypothesis Test of Accuracy between our approach and baselines}
	\label{H1}
	\begin{tabular}{cccccc}
		\toprule
		~           &              & Median ($P_{25},P_{75}$) & Median Difference& Statistic Z values & $p$        \\
		Our\_Acc vs & TPOT         & 0.842(0.6,0.9)   & 0.045                                                 & 2.482                                                  & 0.013*   \\
		& Auto-sklearn & 0.896(0.7,0.9)   & -0.009                                                & 1.83                                                   & 0.067    \\
		& FLAML        & 0.886(0.8,0.9)   & 0.001                                                 & 2.023                                                  & 0.043*   \\
		& mljar        & 0.870(0.7,0.9)   & 0.017                                                 & 3.357                                                  & 0.001**  \\
		\bottomrule
	\end{tabular}
\end{table}
\begin{table}[htbp]
	\centering
	\caption{Hypothesis Test of F1 between our approach and baselines}
	\label{H2}
	\begin{tabular}{cccccc} 
		\toprule
		~          &              & Median ($P_{25},P_{75}$) & Median Difference & Statistic Z values & $p$       \\
		Our\_F1 vs & TPOT         & 0.732(0.4,0.9)   & 0.135                                                 & 3.315                                                  & 0.001**  \\
		& Auto-sklearn & 0.833(0.5,0.9)   & 0.035                                                 & 2.11                                                   & 0.035*   \\
		& FLAML        & 0.810(0.6,0.9)   & 0.058                                                 & 2.386                                                  & 0.017*   \\
		& mljar        & 0.792(0.6,0.9)   & 0.076                                                 & 2.777                                                  & 0.005**  \\
		\bottomrule
	\end{tabular}
\end{table}

\subsubsection{Further Analysis}
To ensure the effectiveness of our method, we do two additional ablation experiments. Since our approach is to enrich the diversity of ensemble strategy, then we fix the ensemble strategy to the stacked generalization approach popular in current AutoML frameworks and compare it. In addition, we compare our approach without enabling the DFP method. The results are shown in Figures 9 and 10.
\begin{figure}[htbp]
	\centering
	\includegraphics[height=6cm,width=14cm]{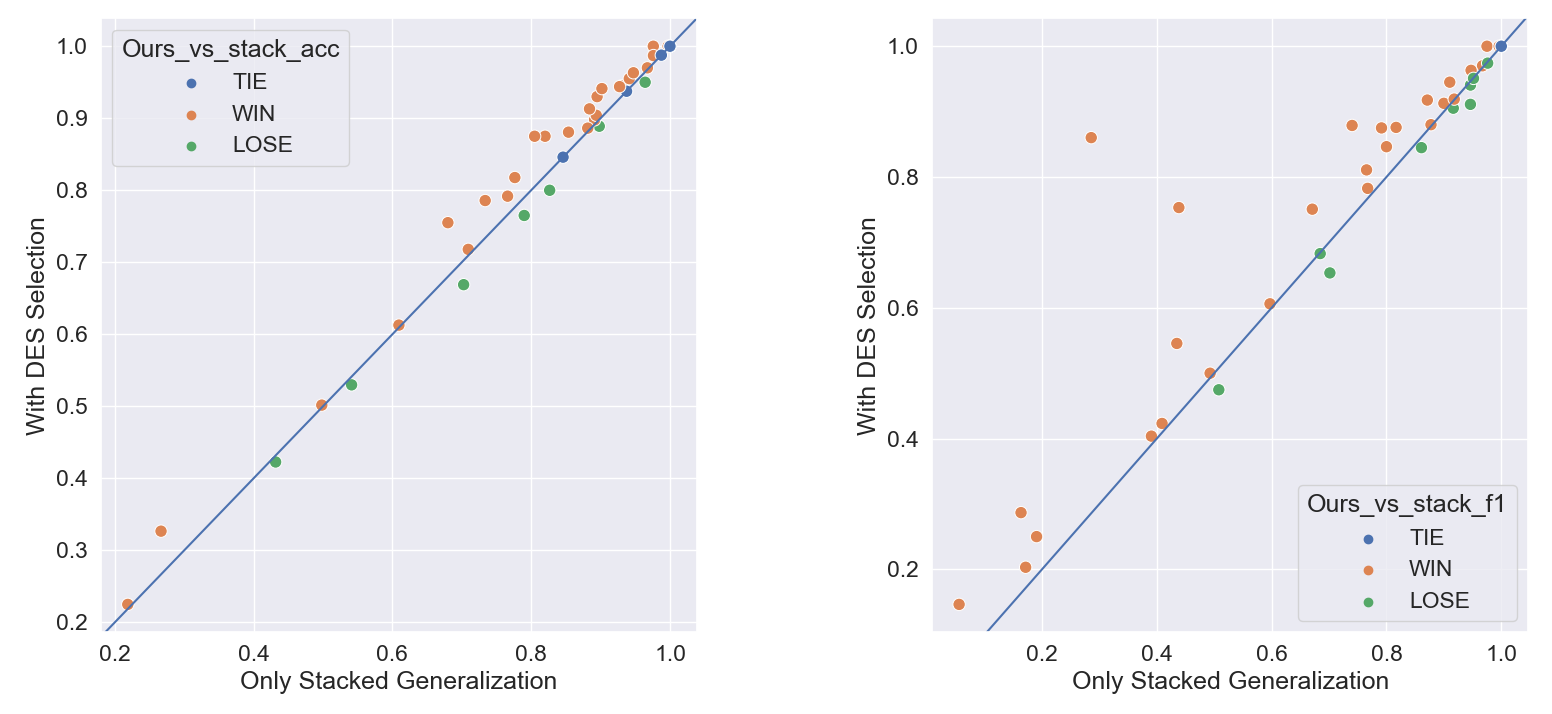}
	\caption{Ablation study of DES selection, the result for accuracy (left) and $F_1$ (right)}
\end{figure}
\begin{figure}[htbp]
	\centering
	\includegraphics[height=6cm,width=14cm]{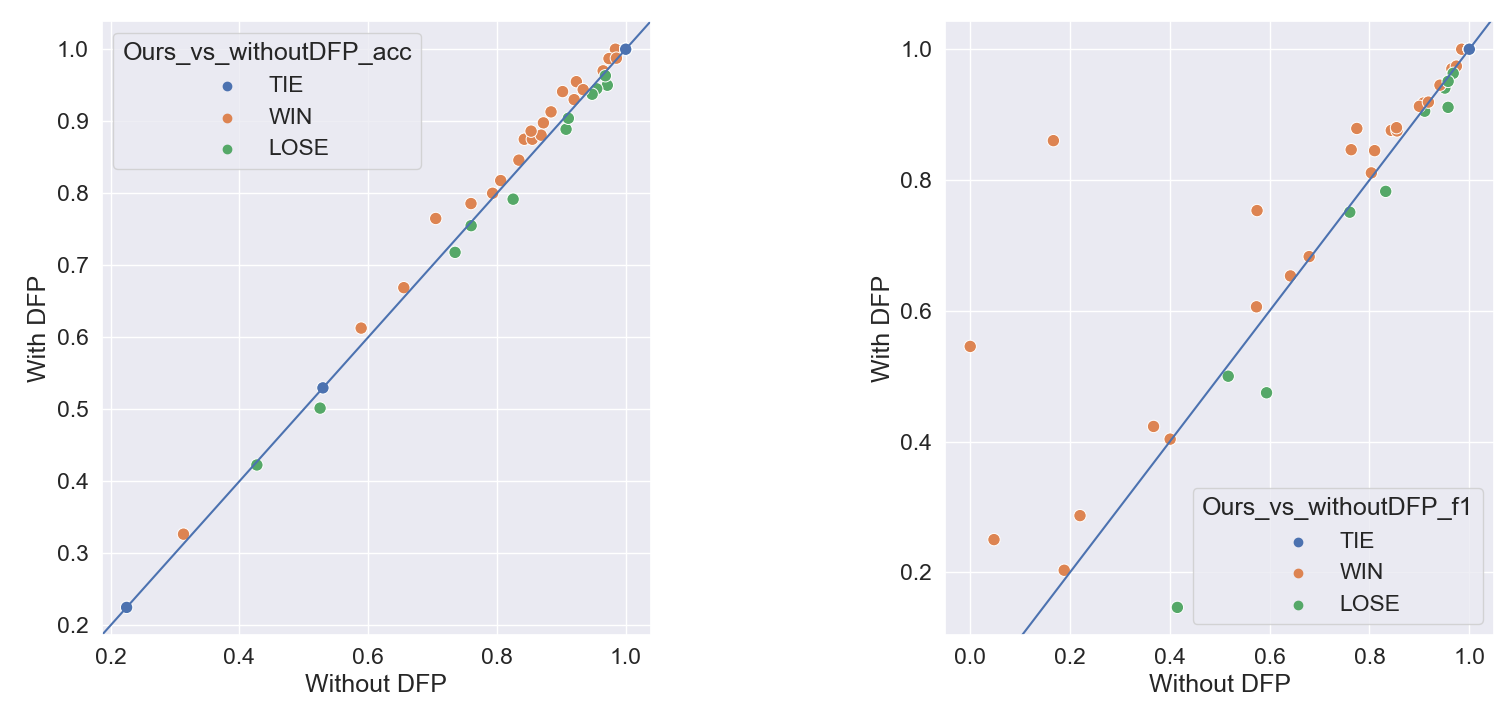}
	\caption{Ablation study of DFP, the result for accuracy (left) and $F_1$ (right)}
\end{figure}
As shown, the vertical axis is the score of the framework of the proposed method, while the horizontal axis is the score of the experiment after ablation. Data points above the line of the $y=x$ function represent better performance before ablation. It can be seen that the module of our proposed ensemble strategy selection method does improve the performance and the DFP method also enhances the performance of the model.

To demonstrate the superiority of our method in dealing with the sample imbalance problem of the dataset, we singled out and compared the datasets with imbalance ratio greater than 2. The results show that our method achieves a more significant advantage in the imbalance datasets.

\begin{table}[htbp]
	\centering
	\caption{Comparison on datasets with imbalance issue}
	\resizebox{\textwidth}{!}{
	\begin{tabular}{cccccc} 
		\toprule
		dataset\_name                    & mljarsupervised\_f1 & TPOT\_F1      & auto-sklearn\_f1  & FLAML\_f1         & AutoDESS\_f1                                          \\
		analcatdata\_authorship          & 0.976421            & 0.976421      & 0.988226          & 0.96429           & \textbf{1}                                            \\
		autoUniv-au6-750        & 0.196211            & 0.268848      & 0.192689          & 0.275965          & \textbf{0.286667}                                     \\
		balance-scale                    & 0.9241              & \textbf{1}    & 0.969123          & 0.960817          & \textbf{1}                                            \\
		blood-transfusion-service-center & 0.230769            & 0.357143      & 0.357143          & 0.357143          & \textbf{0.423077}                                     \\
		cardiotocography                 & \textbf{1}          & \textbf{1}    & \textbf{1}        & \textbf{1}        & \textbf{1}                                            \\
		climate-model-simulation-crashes & 0.941176            & 0.909091      & \textbf{0.951456} & 0.941176          & 0.940594                                              \\
		credit-g                         & 0.801161            & 0.527473      & 0.498084          & 0.843243          & \textbf{0.844884}                                     \\
		eucalyptus                       & 0.607096            & 0.551666      & 0.607241          & 0.613512          & \textbf{0.653288}                                     \\
		ilpd                             & 0.342857            & 0.086957      & \textbf{0.307692} & 0.181818          & 0.146341                                              \\
		kc1                              & 0.347826            & 0.431373      & 0.325581          & 0.425532          & \textbf{0.86019}                                      \\
		pc1                              & 0.4                 & 0.444444      & 0.444444          & 0.25              & \textbf{0.545455}                                     \\
		pc3                              & 0.30303             & \textbf{0.25} & 0.111111          & 0.24              & \textbf{0.25}                                         \\
		pc4                              & 0.622222            & 0.526316      & 0.606061          & \textbf{0.709677} & 0.474576                                              \\
		phoneme                          & 0.833333            & 0.845679      & 0.820513          & 0.855346          & \textbf{0.878816}                                     \\
		steel-plates-fault               & \textbf{1}          & \textbf{1}    & \textbf{1}        & \textbf{1}        & \begin{tabular}[c]{@{}c@{}}\textbf{1}\\\end{tabular}  \\
		volcanoes-a4                     & 0.871393            & 0.87944       & 0.883915          & 0.870751          & \textbf{0.905206}                                     \\
		vowel                            & \textbf{1}          & \textbf{1}    & \textbf{1}        & \textbf{1}        & \textbf{1}                                            \\
		wilt                             & 0.76                & 0.833333      & 0.862745          & 0.791667          & \textbf{0.88}                                         \\
		yeast                            & 0.686872            & 0.686777      & \textbf{0.704862} & 0.68277           & 0.606061                                              \\
		\bottomrule
	\end{tabular}}
\end{table}

The last two figures show our overall statistics for the pipeline generated for all datasets. Figure 11 is a histogram of the ensemble strategies, which shows that a wide variety of strategies were selected for the ensemble. Stacked generalization is indeed a very powerful method resulting in the highest number of selections. Figure 12 shows the heatmap for the combination of classifiers. Each block represents the number of times that combination appears in all pool classifiers. Here we only made a simple visualization that illustrates the method while revealing the intrinsic relationships between models. It also inspires us to improve by performing further data analysis on the generated pipeline or by using meta-learning techniques further.
\begin{figure}[htbp]
	\centering
	\includegraphics[height=6cm,width=14cm]{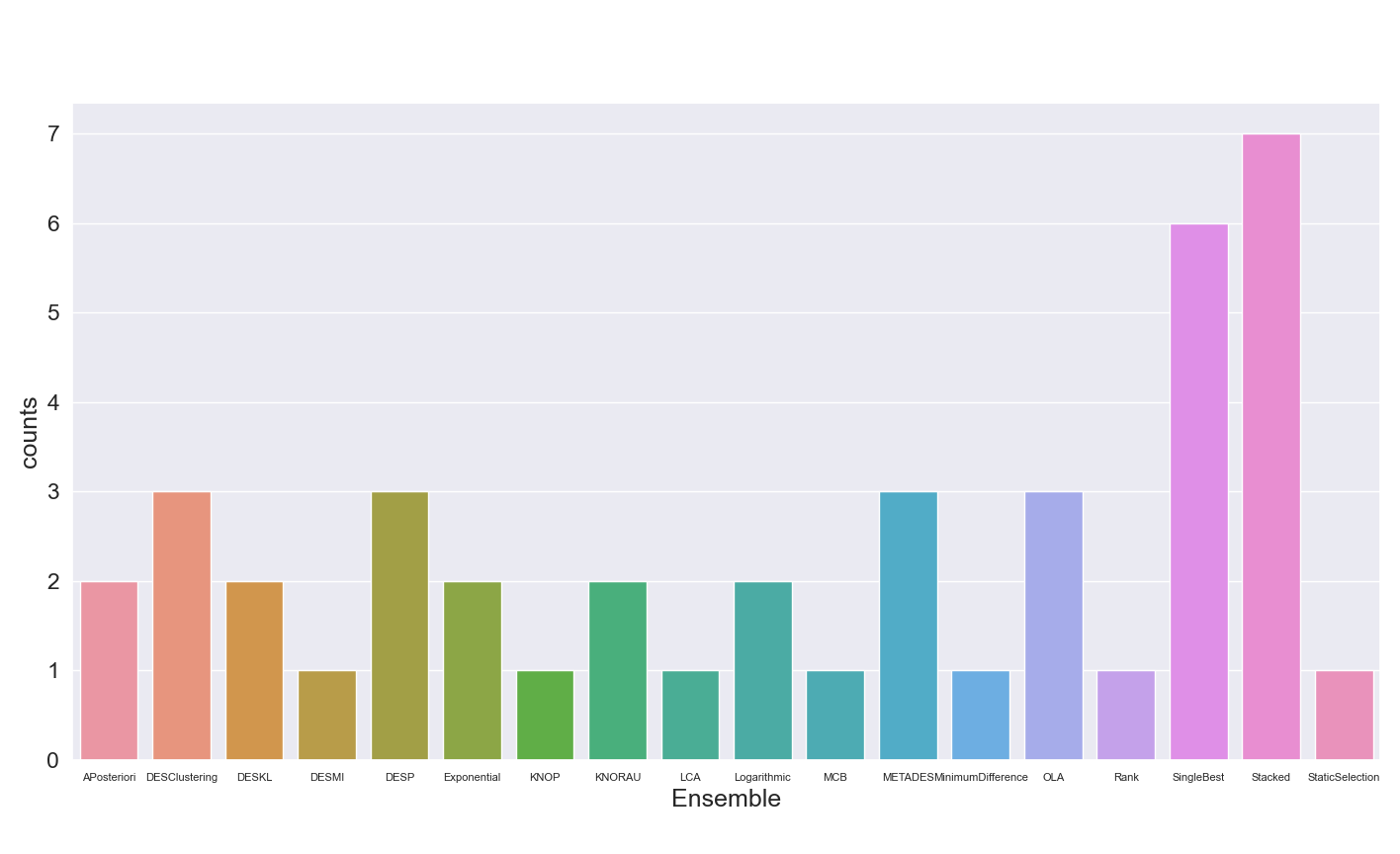}
	\caption{Statistics on ensemble strategies in the generated pipeline}
\end{figure}
\begin{figure}[htbp]
	\centering
	\includegraphics[height=8cm,width=12cm]{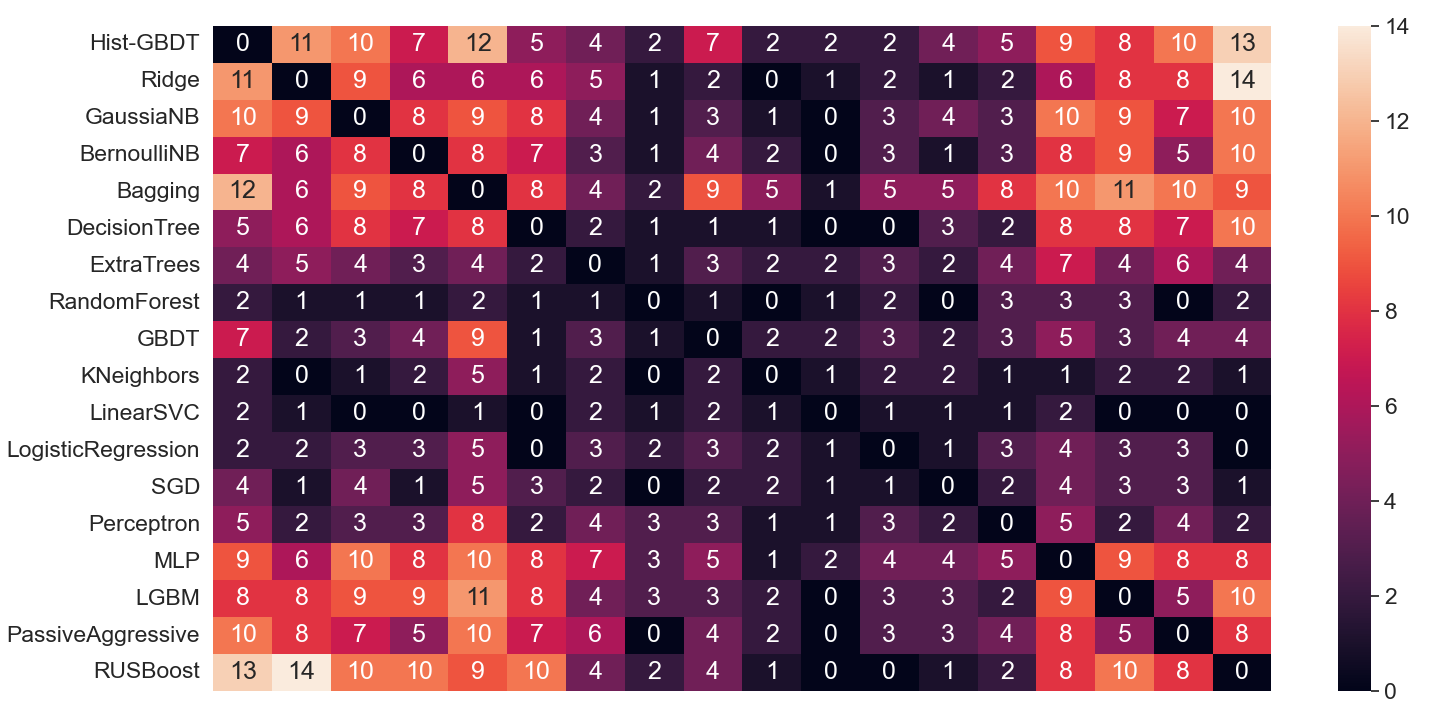}
	\caption{Heatmap of classifier primitives combination}
\end{figure}
\section{Conclusion}
We proposed AutoDESS, an automatic approach of learning ensembles with many different single classifiers with diverse strategies. Compared to 42 different datasets, AutoDESS was able to outperform the current state-of-the-art AutoML for the large majority of the datasets in the same CPU time. Furthermore, we prove that our pro-posed approach is effective in an additional ablation study. 

Our method can be further enhanced in some aspects. In literature\cite{DBLP:conf/icde/LiRBZCZ21}, it is pointed out that some aspects and steps of data preprocessing can have some influence on the model results, however, data preprocessing is not automated in our method. Besides, the parameter search space of primitives can be further increased to achieve better performance.
\bibliography{mybibfile}

\end{document}